\documentclass[letterpaper]{article} 
\usepackage{aaai2026}  
\usepackage{times}  
\usepackage{helvet}  
\usepackage{courier}  
\usepackage[hyphens]{url}  
\usepackage{graphicx} 
\usepackage{natbib}  
\usepackage{caption} 
\usepackage{algorithm}
\usepackage{algorithmic}

\usepackage[table]{xcolor} 
\usepackage{multirow}

\usepackage{newfloat}
\usepackage{listings}
\DeclareCaptionStyle{ruled}{labelfont=normalfont,labelsep=colon,strut=off} 
\floatstyle{ruled}
\newfloat{listing}{tb}{lst}{}
\floatname{listing}{Listing}
\newcommand{\pagefoot}[1]{%
  \insert\footins{\noindent\footnotesize #1\par}%
}

\title{Day-Ahead Electricity Price Forecasting for Volatile Markets \\ Using Foundation Models with Regularization Strategy}
\author {
    Kritchanat Ponyuenyong\textsuperscript{\rm 1, \rm 2 \rm *},
    Pengyu Tu\textsuperscript{\rm 1, \rm 3 \rm *},
    Jia Wei Tan\textsuperscript{\rm 1, \rm 3}\\
    Wei Soon Cheong\textsuperscript{\rm 1},
    Jamie Ng Suat Ling\textsuperscript{\rm 1},
    Lianlian Jiang\textsuperscript{\rm 1}
}
\affiliations {
    \textsuperscript{\rm 1}Institute for Infocomm Research (I\textsuperscript{2}R), Agency for Science, Technology and Research (A*STAR), Singapore\\
    \textsuperscript{\rm 2}College of Computing and Data Science (CCDS), Nanyang Technological University (NTU), Singapore\\
    \textsuperscript{\rm 3}School of Electrical and Electronic Engineering (EEE), Nanyang Technological University (NTU), Singapore\\
    kritchan001@e.ntu.edu.sg, tupe0001@e.ntu.edu.sg, jtan662@e.ntu.edu.sg,\\ caedmoncheong@gmail.com, Jamie@i2r.a-star.edu.sg, Jiang\_Lianlian@i2r.a-star.edu.sg
}

\usepackage{bibentry}

\begin{document}

\maketitle

\begin{abstract}
Electricity price forecasting (EPF) is essential for energy markets stakeholders (e.g. grid operators, energy traders, policymakers) but remains challenging due to the inherent volatility and nonlinearity of price signals. Traditional statistical and deep learning (DL) models often struggle to capture complex temporal dependencies and integrate heterogeneous data effectively. While time series foundation models (TSFMs) have shown strong performance in general time series forecasting tasks, such as traffic forecasting and weather forecasting. However, their effectiveness in day-ahead EPF, particularly in volatile markets, remains underexplored. This paper presents a spike regularization strategy and evaluates a wide range of TSFMs, including Tiny Time Mixers (TTMs), MOIRAI, MOMENT, and TimesFM, against traditional statistical and DL models such as Autoregressive Integrated Moving Average (ARIMA), Long-short Term Memory (LSTM), and Convolutional Neural Network - LSTM (CNN-LSTM) using half-hourly wholesale market data with volatile trends in Singapore. Exogenous factors (e.g. weather and calendar variables) are also incorporated into models where applicable. Results demonstrate that TSFMs consistently outperform traditional approaches, achieving up to 37.4\% improvement in MAPE across various evaluation settings. The findings offer practical guidance for improving forecast accuracy and decision-making in volatile electricity markets.
\end{abstract}


\section{Introduction}

Electricity Price Forecasting (EPF) has become increasingly critical and complex in recent years, driven by the transition towards decentralized energy markets, heightened weather variability and fluctuations in global supply chains. 
Early studies in this field primarily relied on 1) \textit{Statistical models}, such as Autoregressive Integrated Moving Average (ARIMA) \cite{contrerasARIMAModelsPredict2003}, Generalized Autoregressive Conditional Heteroskedasticity (GARCH) \cite{girishSpotElectricityPrice2016}, exponential smoothing methods (ESM) \cite{jonssonForecastingElectricitySpot2013}, general additive models (GAM) \cite{gaillardAdditiveModelsRobust2016} and jump-diffusion models (JDM) \cite{weronModelingElectricityPrices2004}. 
However, each of these models has its limitations. 
For example, ARIMA and GAM capture linear relationship effectively but struggle with nonlinear patterns. 
GARCH performs well only in the presence of relatively high volatility and spikes. 
ESM is simple but inaccurate for rapidly changing data and JDM has difficulty modeling spike clustering \cite{weronElectricityPriceForecasting2014},  2) \textit{Traditional machine learning (ML) and deep learning (DL) models}, such as Random Forests (RF) \cite{meiRandomForestMethod2014}, Extreme Gradient Boosting (XGB) \cite{magalhaesShortTermLoadForecasting2024}, Long-short Term Memory (LSTM) \cite{kongShortTermResidentialLoad2019,marinoBuildingEnergyLoad2016,jiangDayAheadPriceForecasting2018}, Gated Recurrent Units (GRU) \cite{ugurluElectricityPriceForecasting2018}, Bidirectional LSTM (BiLSTM) \cite{chenBRIMAccurateElectricity2019}, PatchTST \cite{nieTimeSeriesWorth2022}, Amplifier \cite{feiAmplifierBringingAttention2025},  and 3) \textit{Hybrid and ensemble models}, such as hybrid Convolutional Neural Network,  CNN-LSTM \cite{huangNovelHybridDeep2021} and hybrid ensemble \cite{bibiElectricitySpotPrices2021}.
These models have demonstrated strong capabilities in modeling temporal dependencies in electricity prices by capturing both local patterns and long-range trends \cite{mubarakDayAheadElectricityPrice2024}. 
However, these approaches usually fail in forecasting complex non-linear price signals.

More recently, time series foundation models (TSFMs) have emerged as a powerful tool for general-purpose forecasting. 
Foundation models such as MOIRAI \cite{wooUnifiedTrainingUniversal2024}, MOMENT \cite{goswamiMOMENTFamilyOpen2024}, TTMs \cite{ekambaramTinyTimeMixers2024}, TimesFM \cite{dasDecoderonlyFoundationModel2024}, and Time-MoE \cite{shiTimeMoEBillionScaleTime2024} are pre-trained on large diverse time series datasets, enabling downstream tasks including forecasting, imputation, classification and anomaly detection in both zero-shot and fine-tuned settings. Studies have shown that TSFMs often match or even surpass traditional methods across a variety of tasks, such as electricity transformer temperature prediction, electricity load forecasting, weather prediction, and exchange trend analysis \cite{montetBenchmarkingFoundationModels2025,ansariChronosLearningLanguage2024}. 
Despite their potential, the application of TSFMs to the EPF domain remains limited. In Singapore, electricity prices are determined through half-hourly real-time bidding and are highly sensitive to global fuel price shocks and geopolitical events due to heavy reliance on imported natural gas. 

Electricity spot prices exhibit seasonality, volatility, and the appearance of price spikes. In the literature on forecasting of electricity price spikes, researchers have noted that static thresholds can misclassify spikes due to inherent volatility and seasonality in electricity prices \cite{repec:eee:appene:v:236:y:2019:i:c:p:196-210}.
Consequently, different variable thresholds have been proposed to better adapt to market dynamics, such as daily or weekly seasonality \cite{repec:gam:jforec:v:6:y:2024:i:1:p:7-137:d:1331777}.
For example, some studies define variable thresholds using mean plus multiple standard deviations of historical prices \cite{2010EPSR...80..318A, vuMultiFeature}, while others use quantile-based limits or autoregressive regime thresholds tailored to specific market features \cite{LU2023106834, LIU2022123417}.
In addition, Kalman filtering techniques have been widely explored for forecasting state variables and prices over time \cite{ofujiPriceForecastingJapan2007}. 
However, price spikes may significantly bias standard Kalman filter estimates, motivating the use of robust Kalman filtering approaches that mitigate the influence of extreme outliers, as demonstrated by \citet{proiettiSpikesMemoryNord2017} in the Nord Pool market.

Combining these insights, in this paper, we propose using TSFMs with a hybrid regularization strategy for day-ahead EPF in Singapore’s volatile high resolution market. The performances of TSFMs based forecasting models are also compared against conventional statistical and DL models.
A standardized experimental framework and consistent evaluation criteria are adopted to ensure fair and reproducible comparisons.
Exogenous influencing factors, such as weather variables and calendar information, are also incorporated into the models as inputs where applicable.
Our findings show that foundation models, particularly TTMs consistently outperform classical methods in both zero-shot and fine-tuned settings, achieving lower forecasting errors and better handling of market volatility.
Moreover, integrating exogenous features and logarithmic transformations further enhances TSFM performance.
These results highlight the potential of foundation models as a promising guideline for future research in EPF, especially in markets characterized by high granularity and dynamic external factors.

\section{Proposed Method}
This section outlines the proposed method framework and key components involved in the preparation and modeling of electricity price data.

\pagefoot{\textsuperscript{1} https://www.nems.emcsg.com/nems-prices\\ \textsuperscript{2} https://www.ema.gov.sg/resources/statistics\\ \textsuperscript{3} https://openweathermap.org/api}

\subsection{Forecasting Framework}
\begin{figure}[!htb]
\centering
\includegraphics[width=1\columnwidth]{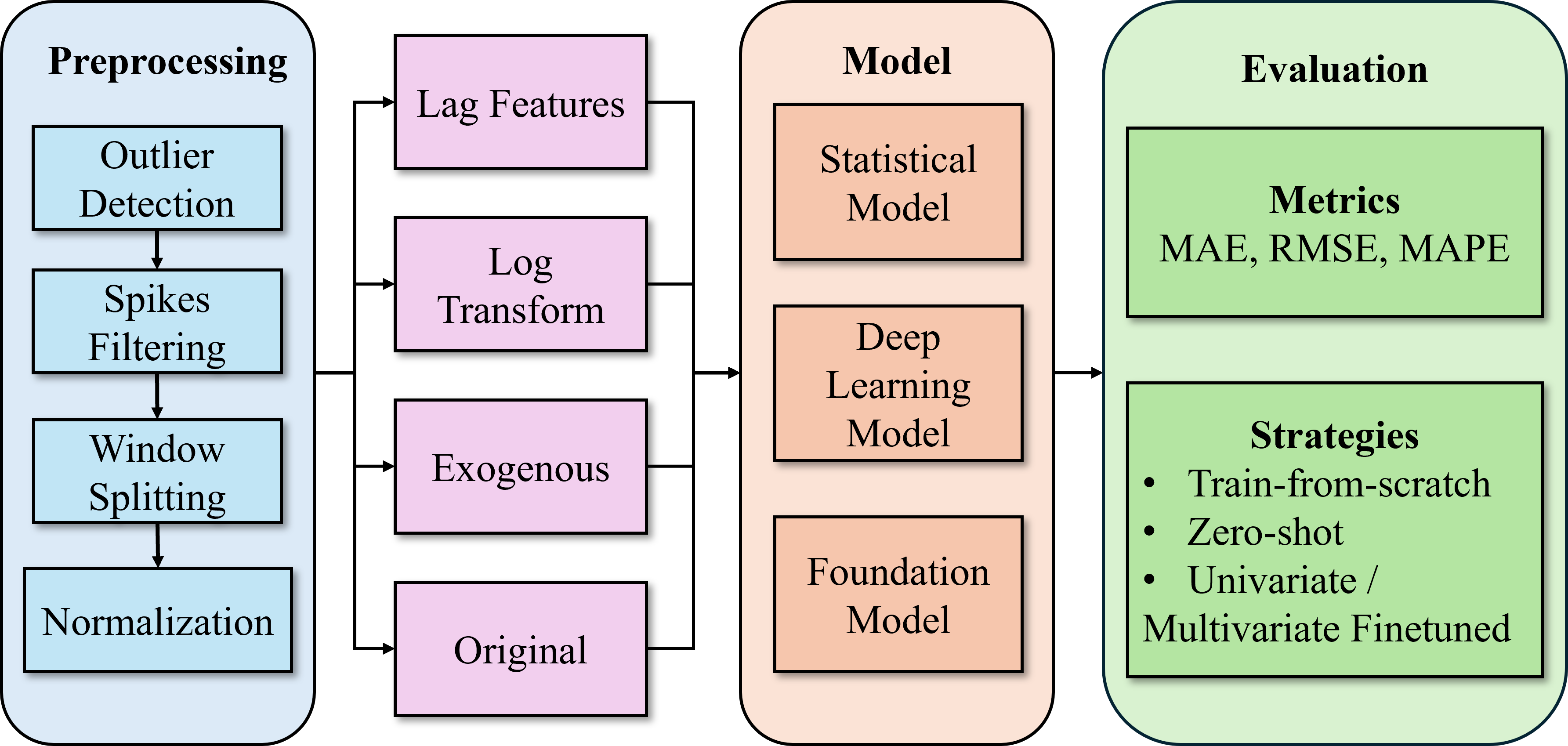} 
\caption{Overall forecasting framework.}
\label{fig1}
\end{figure}

Figure \ref{fig1} illustrates the overall forecasting framework, which consists of four main stages: preprocessing, feature generation, modeling, and evaluation. In the preprocessing stage, raw electricity price data undergoes outlier detection and removal using a seasonal adaptive threshold combined with a robust Kalman filter \cite{kalmanNewApproachLinear1960}. This is followed by window splitting, where the continuous time series is segmented into overlapping windows to create structured input sequences for training. The feature generation stage then derives model inputs by creating lag features, applying log transformations, and incorporating exogenous variables with their corresponding log transformations. All these resulting features, along with the raw inputs, are normalized using the min-max scaling method before being passed to the models. The modeling stage involves training and evaluating of the TSFMs based forecasting models (e.g., TTMs, MOIRAI). As comparison, the statistical models (e.g., ARIMA) and DL models (e.g., LSTM, CNN-LSTM) are also evaluated. Each type of model offers different capabilities in capturing temporal dependencies and complex price patterns. Finally, the evaluation stage assesses model performance using three metrics, including mean absolute error (MAE), root mean square error (RMSE), and mean absolute percentage error (MAPE), under three evaluation strategies: training from scratch (TFS), zero-shot inference with pre-trained models, and supervised fine-tuning. This setup ensures a fair and comprehensive comparison across all tested modeling approaches.

\subsection{Data Input}
The dataset used in this study integrates half-hourly electricity price and demand data from 1 January 2021 to 31 December 2024 obtained from the Energy Market Company\textsuperscript{1} (EMC) and the Energy Market Authority\textsuperscript{2} (EMA) in Singapore, along with weather variables such as temperature, humidity and heat index sourced from the OpenWeather API\textsuperscript{3}. Additionally, public holiday information was accounted for socio-environmental influences on electricity consumption.

The dataset, consisting of 69,600 samples before splitting is divided into training (70\%), validation (20\%), and testing (10\%) sets. In both the training and inferencing stage, the models adopt a sliding window approach, where the time series is divided into overlapping segments. Each segment consists of a fixed lookback window and forecast horizon, with a stride of one. The lookback spans 512-time steps (10 days and 16 hours), and the forecast horizon spans 48 steps (24 hours). This configuration allows the models to learn temporal dependencies without information leakage \cite{yangResearchInformationLeakage2024} and generate multi-step forecasts.

\subsection{Data Preprocessing: Extreme Price Spike Detection and Regularization}
Extreme electricity price spikes, often triggered by transient market shocks or rare events, can introduce substantial volatility and noise into the dataset, potentially distorting model's error function and causing it to overfit to these rare anomalies in the vast majority of normal periods. To address this issue, we proposed a methodology to detect and regularize those extreme price spikes before applying forecasting models to predict the main price trend. The aim is not to produce a smooth or volatility-free signal, but rather to guide the model to prioritize learning the main underlying structure of the data. This allows the model to capture dominant patterns and seasonality while preserving natural market fluctuations, resulting in more stable and reliable forecasts compared to using raw, unfiltered data. 

\begin{figure}[!htb]
\centering
\includegraphics[width=1\columnwidth]{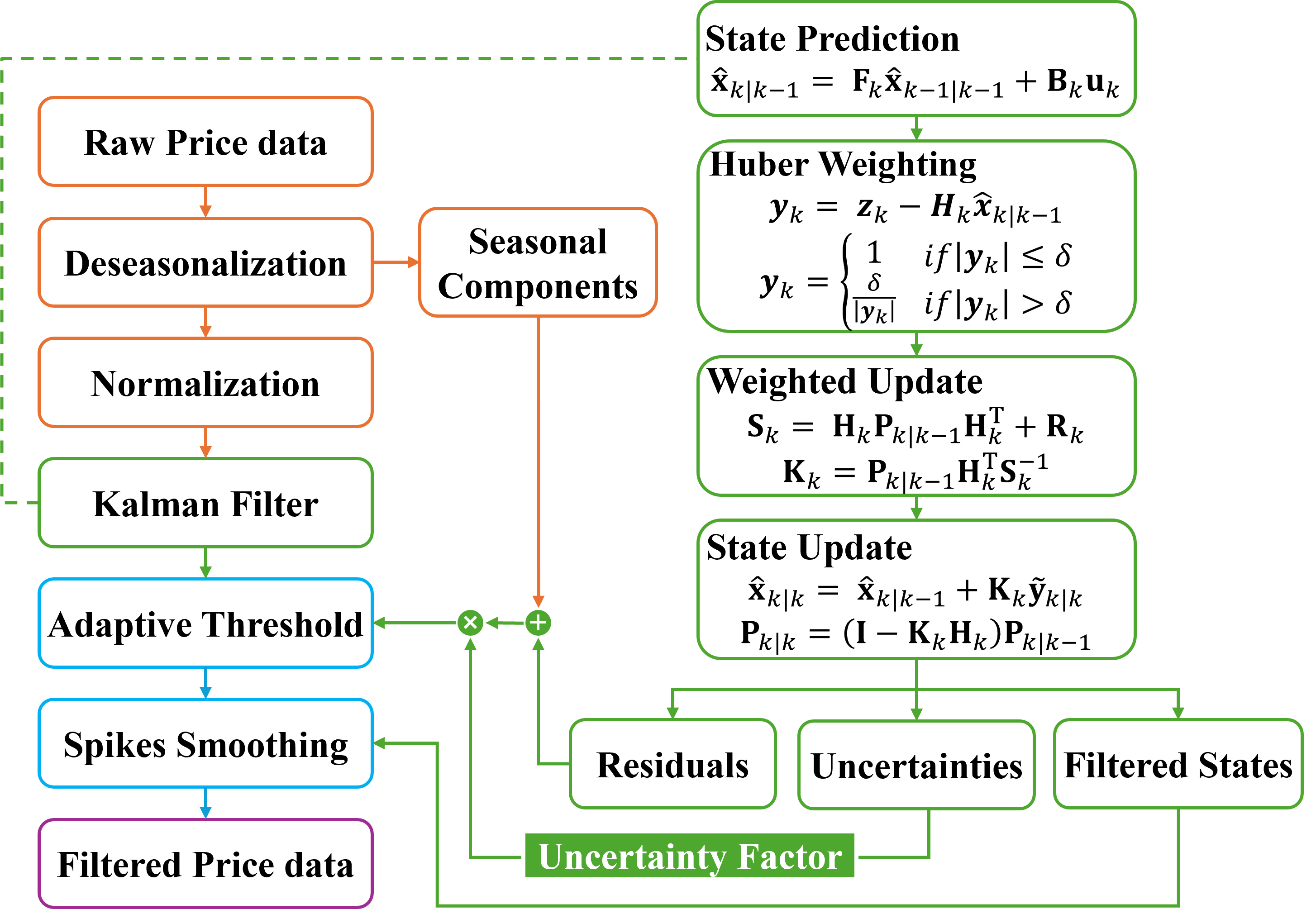}
\caption{Workflow of spikes detection and regularizing strategy STL-KF on raw electricity price data.}
\label{fig2}
\end{figure}

To detect and regularize extreme price points, we developed a three-stage spikes detection and regularization strategy STL-KF as shown in Figure \ref{fig2}: 1) \textit{Season-Trend decomposition using LOESS (STL)} \cite{cleveland1990stl} is first applied to remove cyclical patterns at weekly and monthly levels, thereby isolating non-seasonal deviations in price; 2) a \textit{Kalman Filter (KF)} enhanced with Huber loss functions is then employed to estimate the underlying price states while minimizing sensitivity to outliers; and 3) an \textit{adaptive thresholding mechanism} is used to identify spikes by adding the seasonal components back to the residuals, incorporating an uncertainty factor derived from KF. The corresponding filtered state estimates are extracted and used as replacement values for the identified anomalies.

The filter estimates the state $\hat{x}$ and its error covariance $P$ using a model defined by matrices for transition ($F_k$), observation ($H_k$) and control-input ($B_k$ on vector $u_k$). It incorporates the measurement $z_k$ while accounting for process ($Q_k$) and measurement noise ($R_k$). For robustness, the residual $y_k$ is compared to a threshold $\delta$ to find a weight $w_k$, which modulates the Kalman gain $K_k$. This gain, along with the identity matrix $I$, determines the final state correction. To define the spike detection boundary, we formulate a dynamic confidence interval that accounts for both seasonality and the instantaneous system uncertainty estimated by the Kalman Filter. Let $z_k$ denote the observed price at time step $k$. The adaptive acceptance interval $[LB_k, UB_k]$ is calculated as:

\begin{equation}
    UB_k = (\hat{x}_{k|k-1} + \mathcal{S}_k) + \lambda \cdot \sqrt{P_{k|k-1} + R_k}
\label{eq1}
\end{equation}

\begin{equation}
    LB_k = (\hat{x}_{k|k-1} + \mathcal{S}_k) - \lambda \cdot \sqrt{P_{k|k-1} + R_k}
\label{eq2}
\end{equation}

\noindent where $\hat{x}_{k|k-1}$ is the estimate of the state a priori (the predicted non-seasonal trend) and $\mathcal{S}_k$ denotes the seasonal component derived from the initial STL decomposition. The term $\sqrt{P_{k|k-1} + R_k}$ represents the dynamic uncertainty factor. It aggregates the system error covariance $P_{k|k-1}$ and the measurement noise covariance $R_k$ to adjust the acceptance region based on model confidence. The scaling factor $\lambda$ (set to $\lambda=3$) determines the width of these bounds. Consequently, an observation is flagged as an extreme spike if $z_k \notin [LB_k, UB_k]$. Unlike static thresholding, this mechanism automatically widens the acceptance range during periods of high instability (large $P_{k|k-1}$), reducing false positives during volatile market transitions.

\begin{figure}[!htb]
\centering
\includegraphics[width=1\columnwidth]{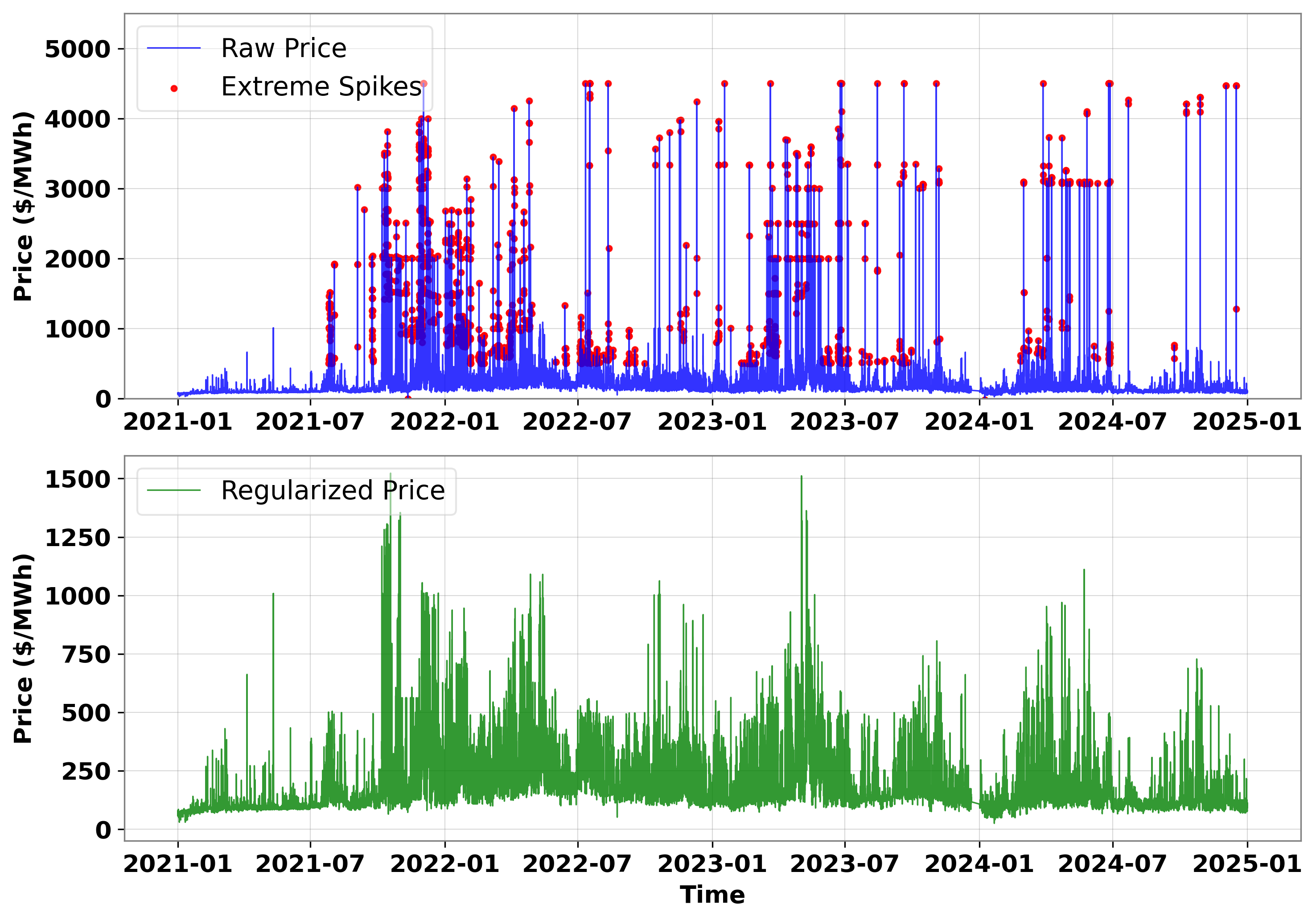}
\caption{Comparison of original electricity price data with detected extreme price spikes in red (top), and the regularized price data after anomaly removal (bottom).}
\label{fig3}
\end{figure}

\begin{figure}[!htb]
\centering
\includegraphics[width=1\columnwidth]{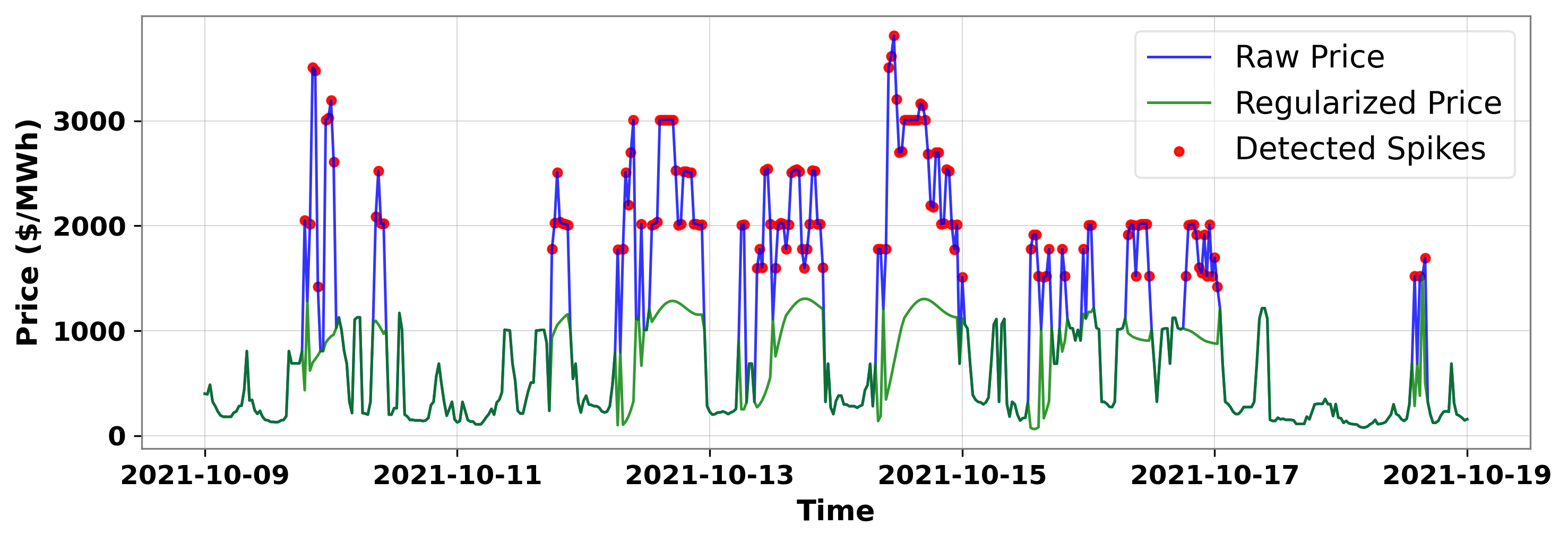}
\caption{A closer view of the identified spikes and regularized signals in a randomly selected window.}
\label{fig4}
\end{figure}

Figure \ref{fig3} shows the original electricity price data (top) with identified extreme price spikes highlighted in red, and the regularized price data after anomaly removal (bottom). The regularized price data exhibits significantly reduced amplitude and volatility while preserving the overall trends. To provide a closer view of the identified spikes and regularized signals, a high-volatility window was randomly selected as shown in Figure \ref{fig4}. Red markers indicate points which are flagged and removed by the proposed method, while the green line represents the smoothed price estimate. The spikes detection strategy can effectively isolate anomalies without over smoothing genuine trend movements.

\subsection{Feature Engineering}
Several feature engineering techniques were employed to enhance the ability of the model to better capture underlying patterns in electricity price data. First, lag features at multiple time steps (\textit{t}-1, \textit{t}-2, \textit{t}-4, \textit{t}-24, \textit{t}-48, \textit{t}-96, \textit{t}-192, \textit{t}-336) were created to capture both short-term dependencies and weekly seasonality in price fluctuations. Second, a natural logarithmic transformation was applied to electricity prices to mitigate the influence of extreme price spikes and stabilize the input distribution. Third, a correlation analysis identified six exogenous variables most strongly associated with electricity prices: demand, hour of the day, temperature, humidity, heat index, and weekend indicators. In total, this resulted in 14 features used for model development.

For the ARIMA model, we employed an auto-ARIMA configuration within a sliding window framework, resulting in varying $p$ and $q$ values for each window. The CNN-LSTM model consists of two convolutional layers and an LSTM block with two hidden layers. The activation function used in the output stage was sigmoid, with mean squared error (MSE) as the loss function. The model was optimized using the Adam optimizer with a learning rate of 0.001. Prior to model training, all features were normalized using min-max scaling to ensure consistent input ranges, computed as: \{ ${x_i^\prime={(x}_i-min(X))/(max(X)-min(X))}$ \}. This normalization prevents features with larger absolute values from dominating the learning process.

\subsection{Model Configurations}
Table 1 summarizes the main types of TSFMs considered along with the compared statistical and DL models with their versions, parameter sizes, architecture types, and references. By considering three types of training strategies (Train-from-scratch, zero-shot, fine-tuning) for the tested TSFMs and compared models, there are 37 models in total. The TSFMs, including MOIRAI, MOMENT, TTMs, TimesFM, Time-MoE, Timer-XL, Lag-Llama and Chronos, are pre-trained on large-scale time series datasets and vary in architecture and parameter sizes. These models are evaluated under both zero-shot and fine-tuned settings. Statistical models such as ARIMA, DL models such as LSTM, CNNLSTM, PatchTST and Amplifier were implemented using standard libraries like Statsmodels and PyTorch, serving as baselines for comparison.

\begin{table}[!htb]
\centering
\caption{Overview of statistical, DL, and TSFMs examined in this study.}
\resizebox{\columnwidth}{!}{%
{\fontsize{14pt}{14pt}\selectfont
\begin{tabular}{
  c@{\hspace{4pt}}
  c@{\hspace{5pt}}
  c@{\hspace{6pt}}
  c@{\hspace{6pt}}
  c@{\hspace{8pt}}
  c@{}
}

\hline
\rule{0pt}{2.2ex}
\textbf{No.} & \textbf{Model} & \textbf{Ver.} & \textbf{Params} & \textbf{Architecture} & \textbf{Reference} \\
\hline
1  & ARIMA       & Auto  & --     & Local             & Weron, 2014 \\
2  & LSTM        & --    & 0.23M  & RNN               & Kong et al., 2019 \\
3  & CNN-LSTM    & --    & 0.32M  & Hybrid CNN        & Mubarak et al., 2024 \\
4  & Amplifier   & --    & 0.34M  & MLP               & Fei et al., 2025 \\
5  & PatchTST    & --    & 3.18M  & Encoder           & Nie et al., 2023 \\
6  & MOIRAI      & Large & 311M   & Encoder           & Woo et al., 2024 \\
7  & MOMENT      & Large & 346M   & Encoder           & Goswami et al., 2024 \\
8  & TTMs        & R2.1  & 0.8M   & MLP-Mixers        & Ekambaram et al., 2024 \\
9  & Time-MoE    & Large & 453M   & MoEs              & Shi et al., 2024 \\
10 & TimesFM     & 500M  & 500M   & Decoder           & Das et al., 2024 \\
11 & Timer-XL    & Base  & 84M    & Decoder           & Liu et al., 2024 \\
12 & Lag-Llama   & Base  & 2.45M  & Decoder           & Rasul et al., 2024 \\
13 & Chronos     & Base  & 205M   & T5                & Ansari et al., 2024 \\
\hline
\end{tabular}
}}
\fontsize{8pt}{8pt}\selectfont
\begin{tabular}{@{}p{0.40\linewidth}p{0.54\linewidth}@{}}
MLP -- Multilayer Perceptron & RNN -- Recurrent Neural Network \\
MoEs -- Mixture of Experts   & T5  -- Text-to-Text Transfer Transformer \\
\end{tabular}
\label{table1}
\end{table}



\section{Results and Discussion}
This section presents the performance of TSFMs based forecasting models and the insights from comparison with traditional statistical and DL models. All models considered use three types of metrics. The comparison results across statistical, deep learning, and foundation models highlight strength of TSFMs based forecasting models.

\subsection{Evaluation Metrics}
Model performance was evaluated using three standard metrics: MAE, MAPE and RMSE. MAE measures the mean absolute error between the forecasted and actual values, given by Eq. \ref{eq3}. MAPE quantifies the average error as a percentage of the actual values (Eq. \ref{eq4}). RMSE, defined in Eq. \ref{eq5}, is the square root of the mean squared error and penalizes larger deviations more heavily. These metrics are defined as:

\begin{equation}
\mathrm{MAE}(y,\hat{y}) = \frac{1}{N} \textstyle\sum\nolimits_{i=1}^{N} \left| y_i - \hat{y}_i \right|,
\label{eq3}
\end{equation}

\begin{equation}
\mathrm{MAPE}(y,\hat{y}) = \frac{1}{N} \textstyle\sum\nolimits_{i=1}^{N} \left| \frac{y_i - \hat{y}_i}{y_i} \right| \times 100\%,
\label{eq4}
\end{equation}

\begin{equation}
\mathrm{RMSE}(y,\hat{y})=\sqrt{\frac{\sum_{i=1}^{N}(y_i-\hat{y}_i)^2}{N}},
\label{eq5}
\end{equation}

\noindent where $N$ represents the total number of forecasted points, $y_i$ and $\widehat{y_i}$ are the actual and predicted values at time step $i$, respectively. MAPE is prioritized as the main evaluation metric in this study as it expresses forecast errors as a percentage of actual values and provides clearer interpretability across varying price levels. This is especially important in volatile electricity markets where relative accuracy is more informative than absolute error.

\subsection{Analysis and Discussion}
Table \ref{table2} presents the performance of TSFMs based forecasting models, with comparison to traditional statistical and DL models (37 models in total) for day-ahead EPF using half-hourly data from the Singapore market. The best performing model for each metric is highlighted in bold and underlined, while the second best is highlighted with an underline. The average metric value for each model category is highlighted in italicized bold. Six forecasting strategies were evaluated, including zero-shot (zs), log-price zero-shot (lzs), univariate fine-tuned (uft), train-from-scratch (tfs), log-price train-from-scratch (ltfs), and multivariate fine-tuned (mft).

\begin{table}[!htb]
\centering
\caption{Performance of TSFMs based forecasting models with comparison to traditional statistical and DL models (37 models in total) for day-ahead electricity price prediction using half-hourly data from the Singapore market.}
\resizebox{\columnwidth}{!}{
\begin{tabular}{llccc}
\hline
\rule{0pt}{2.2ex}
\textbf{Category} & \textbf{Model} & \textbf{MAE} & \textbf{MAPE(\%)} & \textbf{RMSE} \\
\hline
\multirow{3}{*}{Statistical Models} 
& ARIMA & 34.91 & 19.08 & 192.35 \\
& ARIMA (log) & 33.67 & 17.61 & 192.81 \\
& \cellcolor[gray]{0.9}\textbf{\textit{Average Statistical}}
& \cellcolor[gray]{0.9}\textbf{\textit{34.29}}
& \cellcolor[gray]{0.9}\textbf{\textit{18.35}}
& \cellcolor[gray]{0.9}\textbf{\textit{192.58}} \\
\hline
\multirow{9}{*}{DL Models}
& Amplifier (tfs) & 30.79 & 15.40 & 190.89 \\
& Amplifier (ltfs) & 29.41 & 13.77 & 192.18 \\
& PatchTST (tfs) & 31.09 & 14.94 & 192.79 \\
& PatchTST (ltfs) & 30.13 & 13.80 & 192.95 \\
& LSTM (tfs) & 33.26 & 17.76 & 190.96 \\
& LSTM (ltfs) & 31.80 & 16.05 & 191.38 \\
& CNN-LSTM (tfs) & 38.95 & 23.78 & \underline{190.49} \\
& CNN-LSTM (ltfs) & 31.58 & 16.24 & 191.61 \\
& \cellcolor[gray]{0.9}\textbf{\textit{Average DL Models}}
& \cellcolor[gray]{0.9}\textbf{\textit{32.13}}
& \cellcolor[gray]{0.9}\textbf{\textit{16.47}}
& \cellcolor[gray]{0.9}\textbf{\textit{191.66}} \\
\hline
\multirow{28}{*}{TSFMs}
& TTMs (zs) & 28.49 & 12.69 & 192.71 \\
& TTMs (lzs) & 28.42 & 12.62 & 192.61 \\
& TTMs (uft) & \underline{27.96} & \underline{12.41} & 191.85 \\
& TTMs (mft) & \underline{\textbf{27.86}} & \underline{\textbf{11.94}} & 192.19 \\
& Chronos (zs) & 29.14 & 13.28 & 192.99 \\
& Chronos (lzs) & 29.01 & 13.14 & 192.93 \\
& Chronos (uft) & 28.38 & 12.59 & 192.59 \\
& Lag-Llama (zs) & 29.70 & 12.63 & 194.65 \\
& Lag-Llama (lzs) & 29.70 & 12.63 & 194.65 \\
& Lag-Llama (uft) & 30.32 & 13.54 & 194.30 \\
& MOIRAI (zs) & 31.37 & 16.19 & 191.71 \\
& MOIRAI (lzs) & 29.56 & 14.18 & 191.84 \\
& MOIRAI (uft) & 29.37 & 14.04 & 192.51 \\
& MOIRAI (mft) & 29.09 & 13.66 & 191.83 \\
& Timer-XL (zs) & 30.43 & 15.04 & 191.32 \\
& Timer-XL (lzs) & 28.82 & 13.31 & 191.69 \\
& Timer-XL (uft) & 31.15 & 16.27 & \underline{\textbf{189.86}} \\
& Time-MoE (zs) & 30.11 & 14.61 & 191.84 \\
& Time-MoE (lzs) & 29.07 & 13.38 & 192.27 \\
& Time-MoE (uft) & 30.36 & 14.86 & 190.95 \\
& TimesFM (zs) & 29.84 & 13.32 & 193.71 \\
& TimesFM (lzs) & 31.38 & 15.37 & 193.27 \\
& TimesFM (uft) & 30.74 & 15.03 & 192.72 \\
& TimesFM (mft) & 31.40 & 15.76 & 192.48 \\
& MOMENT (zs)\textsuperscript{*} & 61.16 & 38.82 & 236.47 \\
& MOMENT (uft) & 32.23 & 16.62 & 191.97 \\
& MOMENT (mft) & 30.40 & 14.78 & 191.98 \\
& \cellcolor[gray]{0.9}\textbf{\textit{Average TSFMs}}\textsuperscript{*} 
& \cellcolor[gray]{0.9}\textbf{\textit{29.76}}
& \cellcolor[gray]{0.9}\textbf{\textit{13.99}}
& \cellcolor[gray]{0.9}\textbf{\textit{192.47}} \\
\hline
\end{tabular}
}
\fontsize{8pt}{8pt}\selectfont
\begin{tabular}{@{}p{0.46\linewidth}p{0.48\linewidth}@{}}
zs -- Zero–Shot & tfs -- Train from Scratch \\
lzs -- Log Price Zero-Shot & ltfs -- Log Price Train from Scratch \\
uft -- Univariate Fine-tuned & mft -- Multivariate Fine-tuned \\
\end{tabular}
\textsuperscript{*}\,MOMENT (zs) is considered an anomaly thus excluded from average.
\label{table2}
\end{table}

The results show that the multivariate fined-tuned TTMs model - TTMs (mft) achieved the best overall performance in terms of both MAE and MAPE. Notably, all three top-performing models (TTMs (mft), TTMs (uft), Chronos (uft)) are TSFMs under either univariate and multivariate finetuning strategies. Furthermore, across all forecasting settings, foundation models consistently outperformed statistical and DL baseline models, achieving lower average errors (29.76 in MAE and 13.99\% in MAPE) compared to statistical models (34.29 MAE and 18.35\% MAPE) and DL models (32.13 in MAE and 16.47\% in MAPE), as shown in Table \ref{table2}. Unlike DL models, which typically require extensive training and hyperparameter tuning, foundation models achieved superior accuracy even in zero-shot and supervised settings. Recurrent models such as LSTM performed poorly, likely due to their difficulty in capturing the rapid fluctuations characteristics of high-frequency and volatile electricity prices in the Singapore market, despite their strength in modeling long-term dependencies. In contrast, transformer-based models, pretrained on large and diverse time series datasets, demonstrated strong generalization and sequential modeling capabilities. TTMs in particular balance forecasting accuracy and computational efficiency through their simple structure combined with adaptive patching.

\begin{figure}[!htb]
\centering
\includegraphics[width=1\columnwidth]{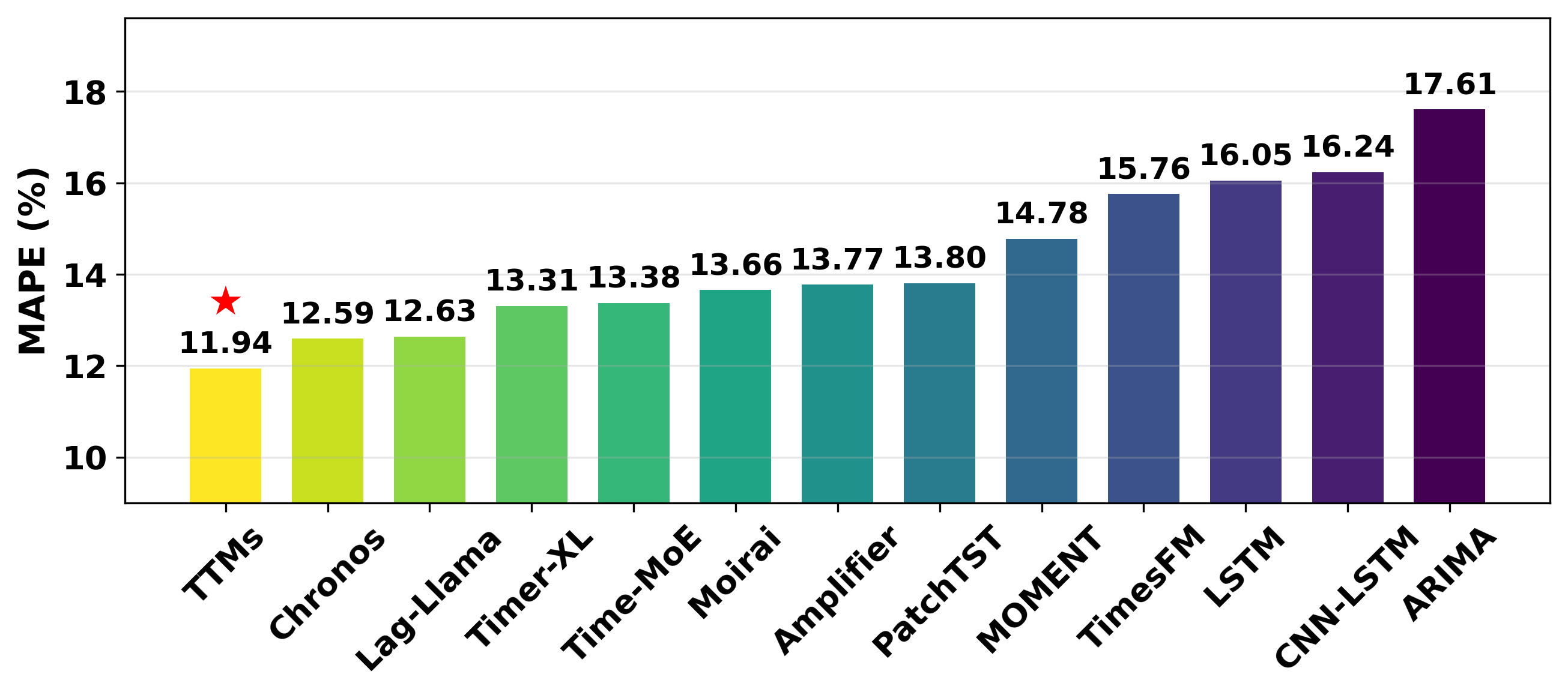}
\caption{Best MAPE performance of each type of model, ranked from lowest to highest error.}
\label{fig5}
\end{figure}

Figure \ref{fig5} ranks all models by their best MAPE performance, highlighting that foundation models such as TTMs, Chronos, Lag-Llama, and Timer-XL consistently outperform traditional statistical and DL models with an improvement of up to 37.4\% in MAPE.

\subsection{Discussion}
This section discusses the impact of preprocessing techniques, zero-shot foundation models and fine-tuning strategies. These insights highlight strengths and opportunities for further improving model robustness.

\subsubsection{Impact of STL-KF based regularization}
\begin{figure}[!htb]
\centering
\includegraphics[width=1\columnwidth]{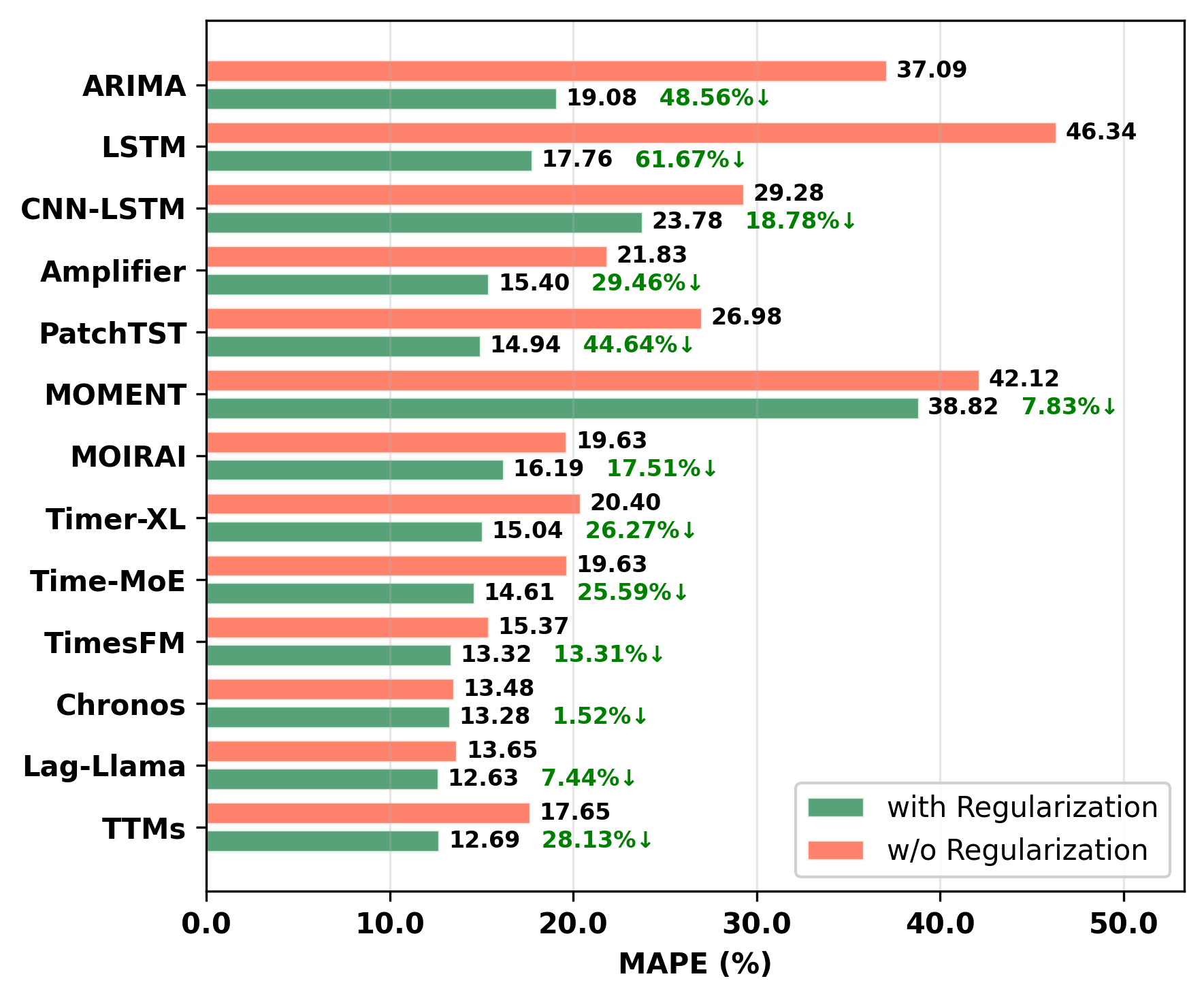}
\caption{MAPE comparison of forecasts using model trained with raw and regularized price data.}
\label{fig6}
\end{figure}

~\\
To evaluate the impact of applying regularization application, the testing dataset, both with and without regularization, was passed through the forecasting models. For foundation models, zero-shot strategy was applied, while statistical and DL models were trained with either raw or regularized data. \textbf{All performance metrics MAPE were computed based on the error between the prediction and original raw price values.} Figure \ref{fig6} compares the MAPE of forecasting results for the best-performing model in each category using raw versus regularized electricity price data. The results show that regularizing noisy market data prior to modeling can significantly improve forecast accuracy. The proposed STL-KF method effectively reduces the impact of price spikes and random noise, achieving up to a 61.67\% (based on LSTM) and an average of 28.69\% improvement in MAPE across all models.

\begin{figure}[!htb]
\centering
\includegraphics[width=1\columnwidth]{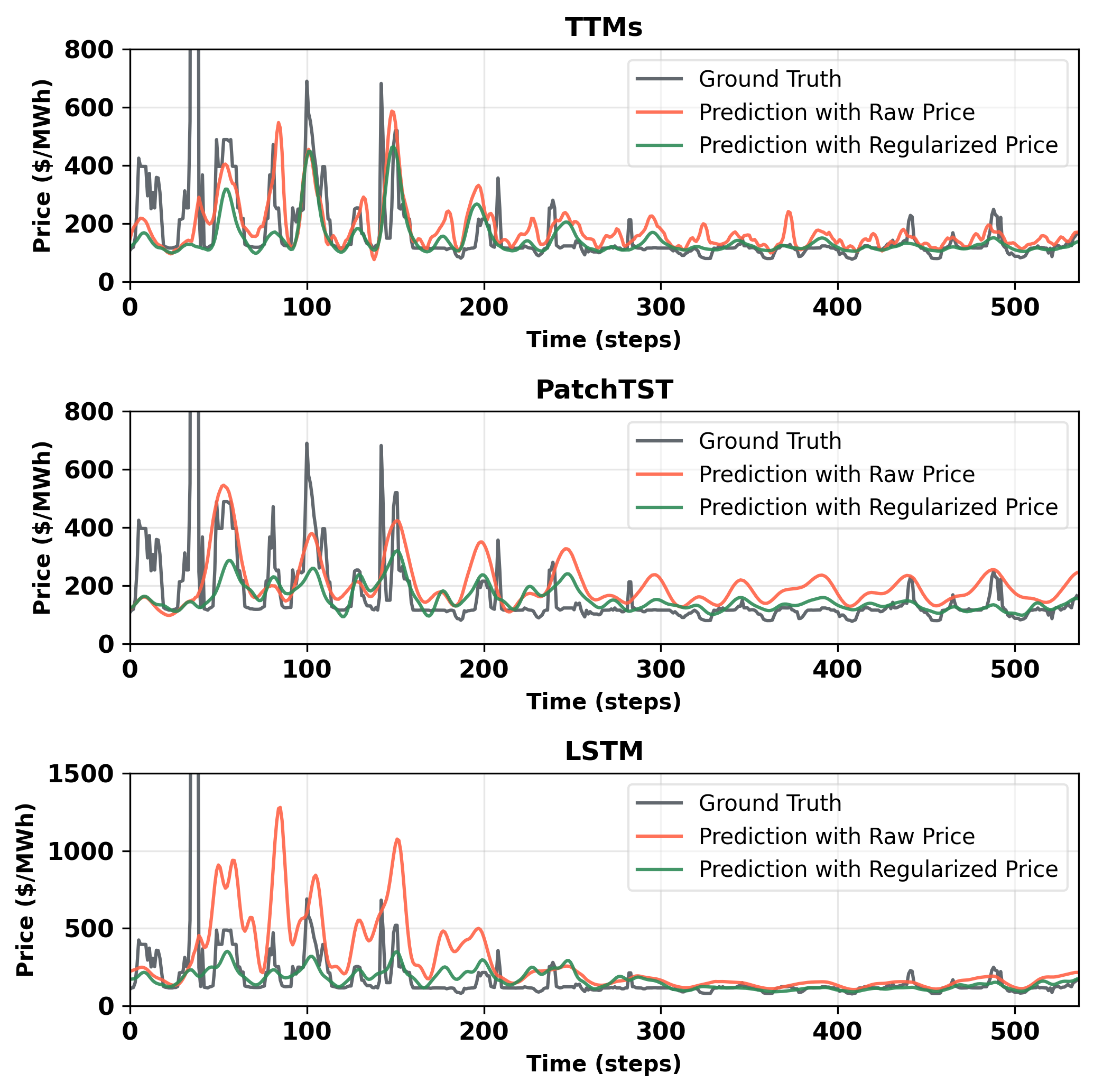}
\caption{Visualization of forecasts (aggregated mean) using regularized and raw data from three types of models.}
\label{fig7}
\end{figure}

Figure \ref{fig7} demonstrates the efficacy of the proposed regularization technique in enhancing the robustness of forecasting models (taking TTMs, PatchTST, and LSTM as examples) against extreme values. While the LSTM model on raw data exhibits high sensitivity, leading to persistently volatile forecasts following an anomaly, its regularized counterpart shows significantly stabilized behavior. Similarly, PatchTST and TTMs also benefit from regularization, achieving improved forecast consistency and a reduced influence from outliers. Collectively, these results confirm that the regularization strategy successfully mitigates the impact of extreme events across model architectures, while preserving the accurate capture of fundamental trend and seasonality.

\subsubsection{Impact of log transformation and features}~\\
Table \ref{table3} shows the impact of log transformation and lag features on the performance of statistical models, DL models and foundation models. This illustrates that the preprocessing technique significantly improves forecasting performance of different model categories. Performing log transformation reduces the MAPE by an average of 9.64\% by smoothing the signal distribution, while adding lag and all other features (e.g. weather data and lag features) provides an additional 3.80\% reduction by capturing patterns at multiple time scales. Log transformations enhance numerical stability, and lag features supply temporal context, benefiting models like LSTM and MOIRAI that rely on sequential memory. In contrast, PatchTST, which already uses convolutional patching, may gain less from lag inputs.

\begin{table}[!htb]
\centering
\caption{Average impact of log transformation and features on MAPE (\%) across model categories.}
\resizebox{\columnwidth}{!}{
\begin{tabular}{lcccc}
\hline
\rule{0pt}{2.2ex}
\textbf{} & \textbf{Statistical} & \textbf{Deep Learning} & \textbf{TSFMs} & \textbf{$\bar{\Delta}$} \\
\hline
w/o Log        & 19.08 & 17.97 & 13.97 & -- \\
w Log          & 17.61 & 14.97 & 13.52 & \textit{\textbf{-9.64\%}} \\
All Features & \textbf{17.09} & \textbf{13.83} & \textbf{13.43} & \textit{\textbf{-3.80\%}} \\
\hline
\end{tabular}
}
\label{table3}
\end{table}





\subsubsection{Comparison of foundation models zero-shot and univariate fine-tuned results}~\\
From Table \ref{table2}, it can be observed that the improvement in accuracy after univariate fine-tuning foundation models is often marginal. In some cases, it decreases the forecasting accuracy. For instance, the MAPE of TTMs is enhanced from 12.62\% in the zero-shot setting to 12.41\% with univariate fine-tuning, showing only a marginal improvement (0.21\%) despite requiring additional training. Chronos shows a MAPE decrease from 13.28\% (zs) to 12.59\% (uft) and MOIRAI from 16.19\% (zs) to 14.04\% (uft). However, for models like Lag-Llama, univariate fine-tuning worsens performance, with MAPE from 12.63\% (zs) to 13.54\% (uft). 
In more extreme cases, models like Timer-XL and TimesFM show significant performance drops after univariate fine-tuning. MAPE of Timer-XL increases from 13.31\% (zs) to 16.27\% (uft), and TimesFM from 13.32\% (zs) to 15.03\% (uft), indicating a clear loss in accuracy. These outcomes suggest that univariate fine-tuning may introduce overfitting by removing the multivariate structure that foundation models are pretrained to leverage. The exclusion of auxiliary inputs such as exogenous variables may limit the ability of models to detect complex dependencies especially in volatile markets.

\subsubsection{Comparison of foundation models zero-shot and multivariate fine-tuned results}~\\
As shown in Table \ref{table2}, the zero-shot evaluation of several pre-trained foundation models such as TTMs, Chronos, Lag-Llama and TimesFM outperformed traditional baselines across key metrics with consistently significantly reduced MAE and MAPE metrics. Among these models, TTMs (mft) emerges as the most stable predictor, achieving the lowest MAE and MAPE across the test set. Meanwhile, hundred million scaled models such as Time-MoE, and Timer-XL demonstrate more responsive behavior to price spikes while still maintaining reliable overall performance achieving RMSE values of 190.95 and 189.86 respectively indicating lower spike errors. These findings suggest that pre-trained temporal representations of foundation models provide robust generalization capabilities for EPF tasks.

\begin{figure}[!htb]
\centering
\includegraphics[width=0.98\columnwidth]{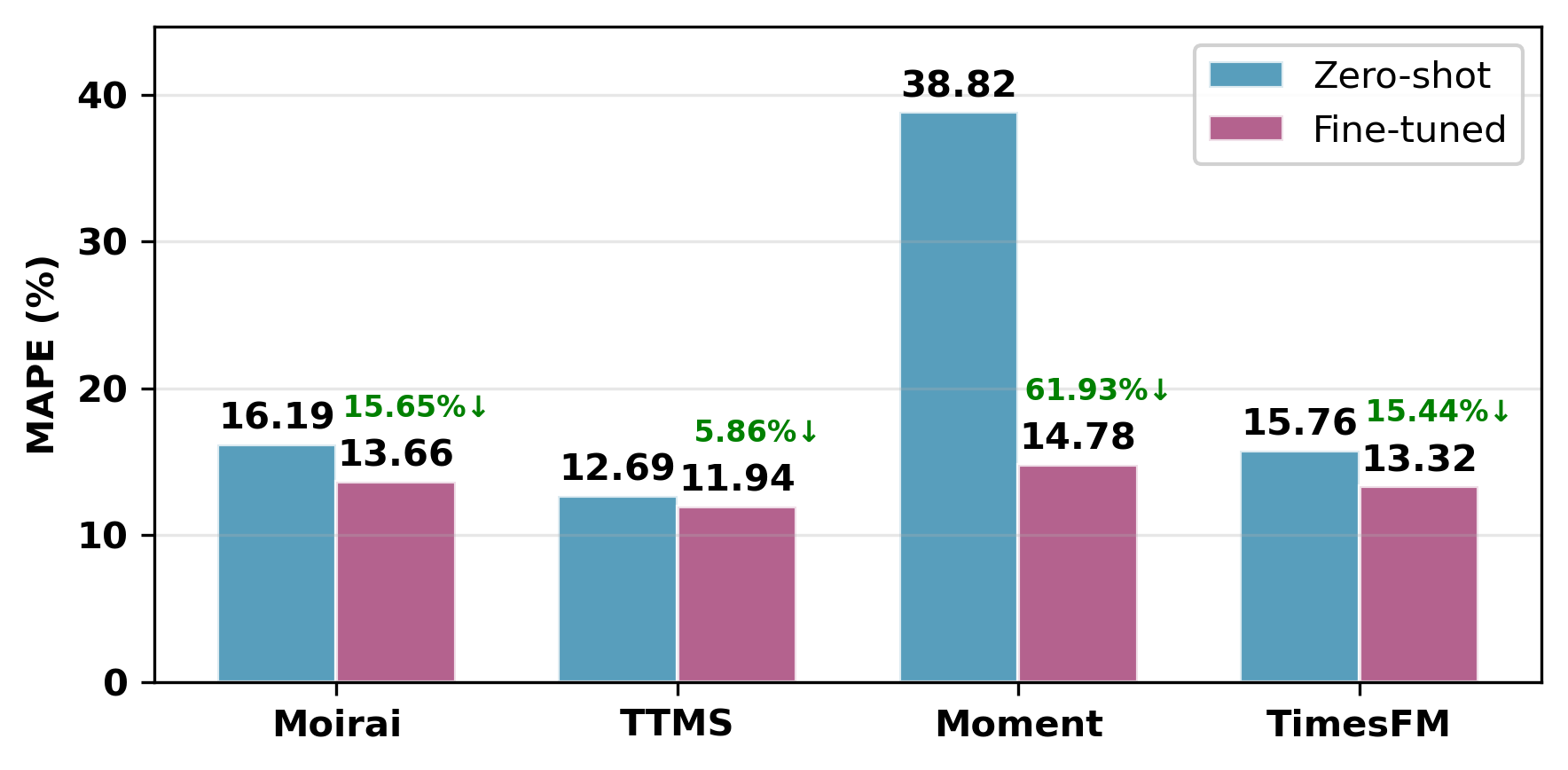}
\caption{MAPE comparison of foundation models zero-shot and multivariate fine-tuned results.}
\label{fig8}
\end{figure}

Figure \ref{fig8} further compares four TSFMs (MOIRAI, MOMENT, TimesFM, and TTMs) fine-tuned on multivariate and zero-shot forecasting on electricity price data. The result highlights the value of incorporating exogenous variables for modeling market dynamics. The selected models were chosen for their compatibility with multivariate inputs and flexible architectures, whereas others were excluded due to constraints like fixed input lengths or lack of dynamic covariate support, which make them less suitable for practical energy forecasting settings.

\subsubsection{Computational Efficiency}~\\
To evaluate practical deployability, we examined the trade-off between architectural copmlexity and inference speed. Transformer-based foundation models like MOIRAI and TimesFM provide strong zero-shot generalization. However, their self-attention layers suffer from quadratic complexity $\mathcal{O}(L^2)$. This causes substantial latency, particularly in autoregressive decoder-only models where multi-step inference is inherently difficult to parallelize. In contrast, LSTM-based methods exhibit linear computational scaling but typically require deep stacks of dilated convolutions to capture the long-range dependencies essential for electricity markets. TTMs offer a balanced alternative for real-time applications. By employing a lightweight MLP-Mixer backbone with adaptive patching, TTMs eliminate the overhead of attention heads. This architecture achieves the global receptive field of Transformers while maintaining low latency and the hardware efficiency of simple matrix operations.

\section{Conclusion}
This paper proposed to use TSFMs for multi-step (day ahead) electricity price forecasting with regularization strategy STL-KF. Eight types of TSFMs, including MOIRAI, MOMENT, TTMs, TimesFM, Time-MoE, Timer-XL, Chronos, and Lag-Llama are evaluated with their comparison to the traditional statistical and deep learning models, such as ARIMA, LSTM, CNN-LSTM and PatchTST. We tested these models specifically on a volatile electricity market such as Singapore. Our results suggest that spike regularization strategy significantly improves the prediction results. TSFMs generally outperform traditional baselines due to pre-trained temporal representations for robust generalization capabilities in EPF tasks. Among these models, TTMs with multivariate finetuning achieve the best performance among the foundation models. Interestingly, applying logarithmic transformation and incorporating lag price and exogenous weather features improved the learning effectiveness of foundation models. Our findings could serve as guidance for users to identify suitable models for their use case. In our future work, we plan to extend our methodology to other markets and investigate the potential of probabilistic forecasting and uncertainty analysis, covering the likelihood of extreme spike events and their quantile ranges.

\bibliography{references}

\end{document}